\def\year{2018}\relax
\documentclass[letterpaper]{article} 
\usepackage{aaai18}  
\usepackage{times}  
\usepackage{helvet}  
\usepackage{courier}  
\usepackage{url}  
\usepackage{graphicx}  
\begin{document}

\title{Developing a Purely Visual Based Obstacle Detection \\ using Inverse Perspective Mapping}
\author{Julian Nubert$^1$, Niklas Funk$^1$, Fabio Meier$^1$, Fabrice Oehler$^1$ \\
{\bf \Large Duckietown Project Report: The Saviors - ETH Zurich} \\ August 2018 \\ \small $^1$\em ETH Zurich, R\"amistrasse 101, 8092 Zurich} 
\nocopyright
\maketitle

\begin{abstract}
Our solution is implemented in and for the frame of Duckietown. The goal of Duckietown\footnote{\url{https://www.duckietown.org/}} is to provide a relatively simple platform to explore, tackle and solve many problems linked to autonomous driving. "Duckietown" is simple in the basics, but an infinitely expandable environment. From controlling single driving Duckiebots until complete fleet management, every scenario is possible and can be put into practice. So far, none of the existing modules was capable of reliably detecting obstacles and reacting to them in real time.

We faced the general problem of detecting obstacles given images from a monocular RGB camera mounted at the front of our Duckiebot and reacting to them properly without crashing or erroneously stopping the Duckiebot. Both, the detection as well as the reaction have to be implemented and have to run on a Raspberry Pi in real time. Due to the strong hardware limitations, we decided to not use any learning algorithms for the obstacle detection part. As it later transpired, a working "hard coded" software needs thorough analysis and understanding of the given problem. 

In layman's terms, we simply seek to make Duckietown a safer place. 
\end{abstract}

\section{Goal}
\label{section:goal}

In practice, a well working obstacle detection is one of the most important parts of an autonomous system to improve the reliability of the outcome in unexpected situations. Therefore the relevance of an obstacle detection in a framework like "Duckietown" is very important, especially since the aim of "Duckietown" is to simulate the real world as realistic as possible. Also in other topics such as fleet planning, a system with obstacle detection behaves completely different than a system without.

\begin{figure}[ht]
  \includegraphics[width=1.0\linewidth]{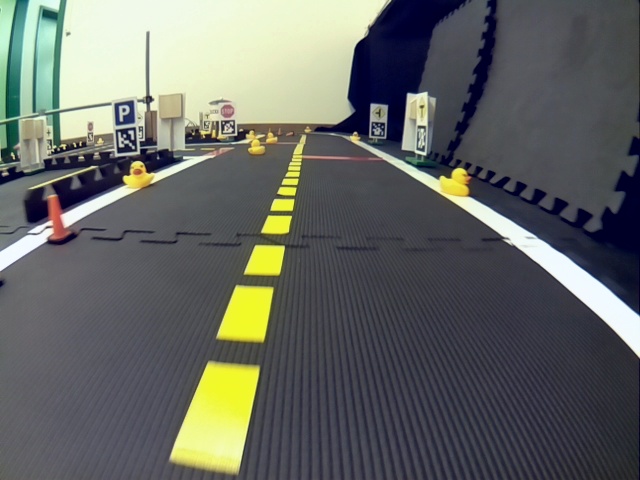}
  \caption{Example Image Taken by the Monocular Camera}
  \label{fig:image_start}
\end{figure}

The goal of our module is to detect obstacles and to react accordingly. Due to the limited amount of time, we focused the scope of our work to the following two points: 
\begin{enumerate}
\item In terms of detection, on the one hand we focused to reliably detect yellow duckies and in this way to save the little duckies that want to cross the road. On the other hand we had to detect orange cones to not crash into any construction sites in Duckietown.
\item In terms of reacting to the detected obstacles we were mainly restricted by the constraint given by the \emph{controllers} of our Duckiebots, who do not allow us to cross the middle of the road. This eliminated the need of also having to implement a Duckiebot detection algorithm. So we focused on writing software which tries to avoid obstacles within our own lane if it is possible (e.g. for avoiding cones on the side of the lane) and to stop otherwise.
\end{enumerate}
The only input into our software pipeline is a RGB colored image, taken by a monocular camera. The input image could look as seen in figure~\ref{fig:image_start}.
With this information given, we want to find out whether an obstacle is in our way or not. If so, we want to either stop or adapt the trajectory to pass without crashing into the obstacle. This information is then forwarded as an output to the \emph{controllers} who will process out commands and try to act accordingly.

One of the first important decisions was to separate the detection (see section~\ref{sec:detection}) and the reaction (see section~\ref{sec:avoidance}) parts of our saviors pipeline. This decision allowed us to divide our work efficiently, to start right away and is supposed to ensure a wide range of flexibility in the future by making it possible to easily replace, optimize or work on one of the parts separately (either the obstacle avoidance strategies or obstacle detection algorithms). Of course it also includes having to define a clear, reasonable interface in between the two modules, which will later be explained in detail.

The whole project software was implemented in \emph{Python} using \emph{ROS (Robot Operating System)}.

\section{Detection - Computer Vision}
\label{sec:detection}

\subsection{Description}

Using the camera image only we wanted to reach the following:
\begin{itemize}
\item \textbf{Detect} the obstacles in the camera image,
\item give \textbf{the 3D coordinates} and \textbf{the size} of every detected obstacle in the real world and
\item label each obstacle if it's \textbf{inside or outside the lane boundaries} (e.g. for the purpose of not stopping in a curve).
\end{itemize}
For code optimizing reasons it was also very important to visualize them in the camera image as well as to plot markers in the 3D world in \textit{rviz}.

\subsubsection{Assumptions}

Since every algorithm has its limitations, we made the following two assumptions:
\begin{itemize}
\item Obstacles are only yellow duckies and orange cones and
\item we use a calibrated camera including \emph{intrinsics} and \emph{extrinsics}.
\end{itemize}

\subsubsection{Robustness}

From the very beginning we clearly formulated the goal to reach a maximum in robustness with respect to changes in:
\begin{itemize}
\item Obstacle Size,
\item obstacle color (within orange and yellow to detect different traffic cones and duckies) and
\item illumination.
\end{itemize}

\subsubsection{Performance Metrics}

For evaluating the performance, we used the following metrics, evaluated under different light conditions and different velocities (static and in motion):
\begin{itemize}
\item Percentage of correctly classified obstacles on our picture data sets,
\item percentage of false positives and
\item percentage of missed obstacles.
\end{itemize}
An evaluation of our goals and the reached performance can be found in the Performance Evaluation section (\ref{sec:perfEval}).

\subsection{Functionality}

Let's again have a look on the usual incoming camera picture in figure~\ref{fig:image_start}. In the very beginning of the project, like the previous implementation in 2016, we tried to do the detection in the normal camera image but we tried to optimize for more efficient and general obstacle descriptors. Due to the specifications of a normal camera, lines which are parallel in the real world are in general not parallel any longer and so the size and shape of the obstacles are disturbed (elements of the same size appear also larger in the front than in the back). This made it very difficult to reliably differentiate between yellow ducks and line segments. We tried several different approaches to overcome this problem, namely:
\begin{itemize}
\item Patch matching of duckies viewed from different directions.
\item Patch matching with some kind of an ellipse (because line segments are supposed to be square).
\item Measuring the maximal diameter.
\item Comparing the height and the width of the objects.
\item Taking the pixel volume of the duckies.
\end{itemize}
Unfortunately none of the described approaches provided a sufficient performance. Also a combination of them didn't make the desired impact. All metrics which are somehow associated with \textbf{the size} of the object just won't work because duckies further away from the duckiebot are simply a lot smaller than the one very close to the Duckiebot. All metrics associated with \textbf{the "squareness"} of the lines were strongly disturbed by the occurring motion blur. This makes finding a general criterion very difficult and made us think about changing the approach entirely.

Therefore we developed and came up with the following new approach.

\subsubsection{Theoretical Description}

\begin{figure}[ht]
  \includegraphics[width=1.0\linewidth]{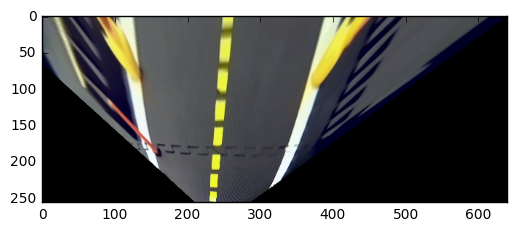}
  \caption{Image seen from Bird's View Perspective}
  \label{fig:bird_view}
\end{figure}

In our setup, through the extrinsic camera calibration, we are given a mapping from each pixel in the camera frame to a corresponding real world coordinate. It is important to mention that this transformation assumes all seen pixels in the camera frame to lie in one plane which is in our case the ground plane/street. Our approach exactly exploits this fact by transforming the given camera image into a new, bird's view perspective which basically shows one and the same scene from above. Therefore the information provided by the extrinsic calibration is essential for our algorithm to work properly. In figure~\ref{fig:bird_view} you can see the newly warped image seen from the bird's view perspective. This is one of the most important steps in our algorithm.
This approach has already been shown by Prof. Davide Scaramuzza (UZH) and some other papers and is referred as \textbf{Inverse Perspective Mapping Algorithm}. (see  \cite{Scaramuzz2005}, \cite{GangYiJiang2000}, \cite{MassimoBertozzi1997}).

What stands out, is that the lines which are parallel in the real world are also parallel in this view. Generally in this "bird's" view, all objects which really belong to the ground plane are represented by their real shape (e.g. the line segments are exact rectangles) while all the objects which are not on the ground plane (namely our obstacles) are heavily distorted in this top view. This top view is roughly keeping the size of the elements on the ground whereas the obstacles are displayed a lot larger.

The theory behind the calculations and why the objects are so heavily distorted can be found in the Theory Chapter in section~\ref{sec:theory}.

Either way, we take advantage of this property. Given this bird's view perspective, we still have to extract the obstacles from it. To achieve this extraction, we first filter out everything except for orange and yellow elements, since we assumed that we only want to detect yellow duckies and orange cones. To simplify this step significantly, we transform the obtained color corrected images (provided by the Anti Instagram module) to the \textbf{HSV color space}. We use this \emph{HSV} color space and not the \emph{RGB} space because it is much easier to account for slightly different illuminations - which of course still exist since the performance of the color correction is logically not perfect - in the \emph{HSV} room compared to \emph{RGB}. For the theory behind the \emph{HSV} space, please also refer to our appropriate Theory Chapter in section~\ref{sec:theory}.

After this first color filtering process, there are only objects remaining which have approximately the colors of the expected obstacles. For the purpose of filtering out the real obstacles from the bunch of all the remaining objects which passed the color filter, we decided to do the following: We segment the image of the remaining objects, i.e. all connected pixels in the filtered image are getting the same label such that you can later analyse the objects one by one. Each number then represents an obstacle. For the process of segmentation, we used the algorithm described by \cite{Wu2005}.

Given the isolated objects, the task remains to finally decide which objects are considered obstacles and which not. In a first stage, there is a filter criterion based on a rotation invariant feature, namely the two eigenvalues of the \emph{inertia tensor} of the segmented region when rotating around its center of mass (see \cite{Fitzpatrick2011}).

In a second stage, we apply a tracking algorithm to reject the remaining outliers and decrease the likelihood for missclassifications. The tracker especially aims for objects which passed the first stage's criterion by a small margin.

\begin{figure}[ht]
  \includegraphics[width=1.0\linewidth]{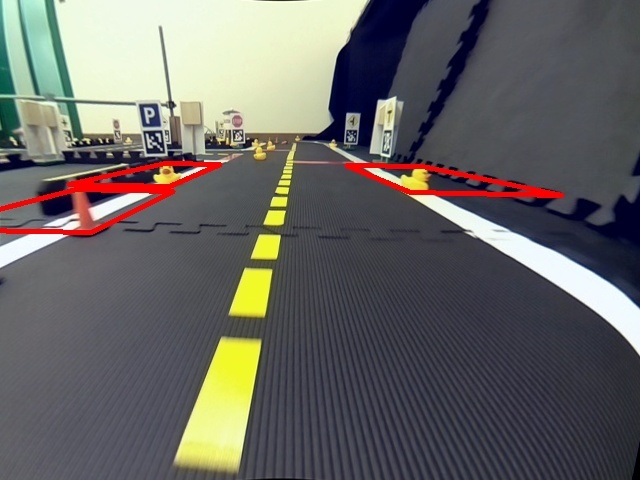}
  \caption{Final Output of the Detection}
  \label{fig:part_1_image_final}
\end{figure}

For further information and details about how we perform the needed operations, please refer to the next chapter. The final output of the detection module is the one which can be seen in figure~\ref{fig:part_1_image_final}.

\subsubsection{Actual Implementation}

Now we want to go more into detail on how we implemented the described steps.

In the beginning we again start from the picture you can see in figure~\ref{fig:image_start}. In our case this is now the corrected image coming out from the image\_transformer\_node and was implemented by the anti instagram group. We then perform the following steps:
\begin{enumerate}
\item In a first step we crop this picture to make our algorithm a little bit more efficient and due to our limited velocities, it makes no sense to detect obstacles which are not needed to be taken into consideration by our obstacle avoidance module. However, we do not simply crop the picture by a fixed amount of pixels, but we use the extrinsic calibration to neglect all the pixels which are farther away than a user defined threshold, which is at the moment at 1.7 meters. So the amount of pixels which are neglected are different for every Duckiebot and depend on the extrinsic calibration. The resulting image can be seen in figure~\ref{fig:image_cropped}. The calculations to find out where you have to cut the image are quite simple (note that it still bargains for homogeneous coordinates): 
\begin{equation}
P_{camera} = H^{-1}P_{world}
\end{equation}
A cutted example image can be found in figure~\ref{fig:image_cropped}.
\begin{figure}[ht]
  \includegraphics[width=1.0\linewidth]{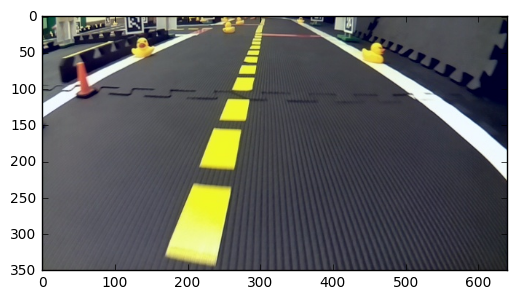}
  \caption{Cropped Image}
  \label{fig:image_cropped}
\end{figure}
\item Directly detecting the obstacles from this cropped input image failed for us due to the reasons described above. That is why the second step is to perform the transformation to the bird’s view perspective. For transforming the image, we first use the corners of the cropped image and transform it to the real world. Then we scale the real world coordinates to pixel coordinates, so that it will have a width of 640 pixels afterwards. For warping all of the remaining pixels with low artifacts we use the function \emph{cv2.getPerspectiveTransform()}. The obtained image can be seen in figure~\ref{fig:bird_view}.
\item Then we transform the given \emph{RGB} picture into the \emph{HSV} colorspace and apply the yellow and orange filter. While a \emph{HSV} image is hardly readable for humans, it is way better to filter for specific colors. The obtained pictures can be seen in figure~\ref{fig:yellow_filtered} and figure~\ref{fig:orange_filtered}. The color filter operation is performed by the \emph{cv2} function \emph{cv2.inRange(im\_test, self.lower\_yellow, self.upper\_yellow)} where lower\_yellow and upper\_yellow are the thresholds for yellow in the \emph{HSV} color space.
\begin{figure}[ht]
  \includegraphics[width=1.0\linewidth]{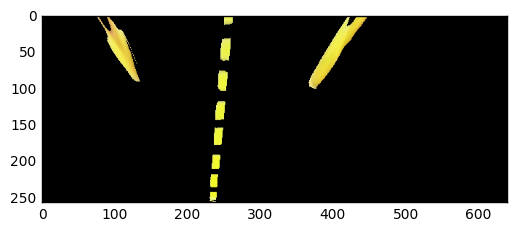}
  \caption{Yellow Filtered Image}
  \label{fig:yellow_filtered}
\end{figure}
\begin{figure}[ht]
  \includegraphics[width=1.0\linewidth]{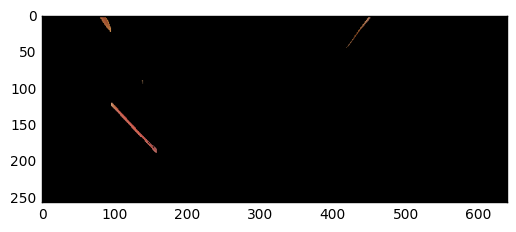}
  \caption{Orange Filtered Image}
  \label{fig:orange_filtered}
\end{figure}
\item Now there is the task of segmenting/isolating the objects which remained after the color filtering process. At the beginning of the project we therefore implemented our own segmentation algorithm which was however more inefficient and led to an overall computational load of 200\% CPU usage and a maximum frequency of our whole module of about 0.5 Hz only. By using the \emph{scikit-image} module which provides a very efficient label function, the computational efficiency could be shrunk considerably to about 70\% CPU usage and allows the whole module to run at up to 3 Hz. It is important to remember that in our implementation the segmentation process is the one which consumes the most computational resources. The output after the segmentation is the one seen in figure~\ref{fig:segmented}, where the different colors represent the different segmented objects.
\begin{figure}[ht]
  \includegraphics[width=1.0\linewidth]{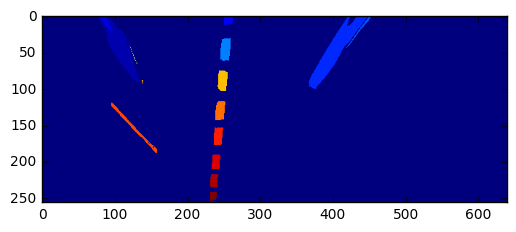}
  \caption{Segmented Image}
  \label{fig:segmented}
\end{figure}
\item After the segmentation, we analyze each of the objects separately. At first there is a general filter which ensures that we are neglecting all the objects which contain less than a user influenced threshold of pixels. As mentioned above, the homographies of all the users are different and therefore the exact amount of pixels an object is required to have must again be scaled by the individual homography. This is followed by a more detailed analysis which is color dependent. On the one hand there is the challenge to detect the orange cones reliably. Speaking about cones, the only other object that might be erroneously detected as orange are the stop lines. Of course, in general the goal should be to very reliably detect orange but as the light is about to change during the drive, we prepared to also detect the stop lines and being able to cope with them when they are erroneously detected. The other general challenge was that all objects that we have to detect can appear in all different orientations. Looking at figure~\ref{fig:bird_view_box}, simply inferring the height and width of the segmented box, as we did it in the beginning, is obviously not a very good measure (e.g. in in the lower left the segmented box is square while the cone itself is not quadratic at all).
\begin{figure}[ht]
  \includegraphics[width=1.0\linewidth]{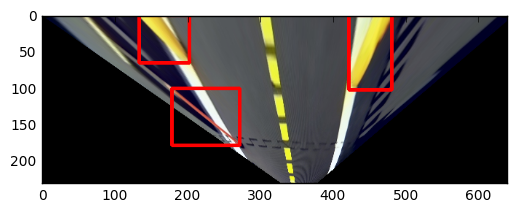}
  \caption{Bird's View with Displayed Obstacle Boxes}
  \label{fig:bird_view_box}
\end{figure}
That is why it is best to use a rotation invariant feature to classify the segmented object. In our final implementation we came up with using the two eigenvalues of the \emph{inertia tensor}, which are obviously rotational invariant (when being ordered by their size). Being more specific about the detection of cones, when extracting the cone from figure~\ref{fig:segmented} it is looking like in figure~\ref{fig:segmented_cone}, while an erroneous detection of a stop line is looking like in figure~\ref{fig:segmented_stop_line}.
\begin{figure}[ht]
  \includegraphics[width=1.0\linewidth]{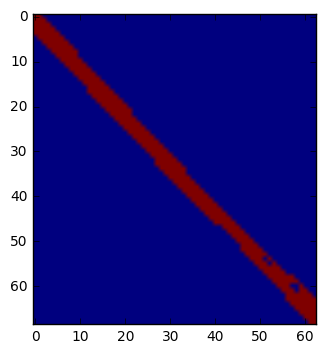}
  \caption{Segmented Cone}
  \label{fig:segmented_cone}
\end{figure}
\begin{figure}[ht]
  \includegraphics[width=1.0\linewidth]{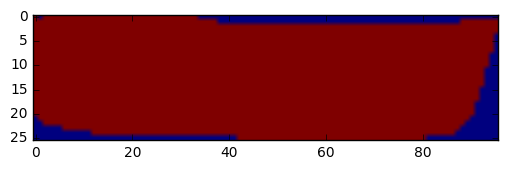}
  \caption{Segmented Stop Line}
  \label{fig:segmented_stop_line}
\end{figure}
Our filter criteria is now the ratio between the eigenvalues of the inertia tensor. This ratio is always by a factor of about 100 greater in case the object is a cone, compared to when we erroneously segment a red stop line. This criteria is very stable that is why there is no additional filtering needed to detect the cones.

If the segmented object is yellowish, things get a little more tricky as there are always many yellow objects in the picture, namely the middle lines. Line elements can be again observed under every possible orientation. Therefore the eigenvalues of the \emph{inertia tensor}, which are as mentioned above rotational invariant, are again the way to go. In figure~\ref{fig:segmented_middle_line} you can see a segmented line element and in figure~\ref{fig:segmented_duck} again a segmented duckie.
\begin{figure}[ht]
  \includegraphics[width=1.0\linewidth]{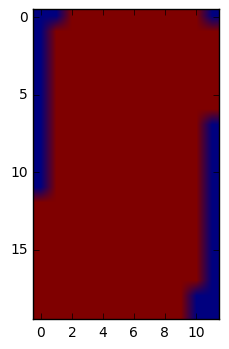}
  \caption{Segmented Middle Line}
  \label{fig:segmented_middle_line}
\end{figure}
\begin{figure}[ht]
  \includegraphics[width=1.0\linewidth]{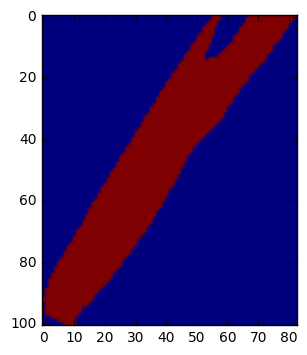}
  \caption{Segmented Duckie}
  \label{fig:segmented_duck}
\end{figure}
As the labeled axis already reveal, they are of a different scale, but as we also got very small duckies, we had to choose a very small threshold. To detect the yellow duckies, the initial condition is that the first eigenvalue has to be greater than 20. This criteria alone however includes to sometimes erroneously detecting the lines as obstacles, that is why we implemented an additional tracking algorithm which works as follows: If an object’s first eigenvalue is greater than 100 pixels and it is detected twice - meaning in two consecutive images there is an object detected at roughly the same place - it is labeled as an obstacle. However, if an object is smaller or changed the size by more than 50\% in the consecutive frames, then a more restrictive criteria is enforced. This more restrictive criterion states that we must have tracked this object for at least for 3 consecutive frames before being labeled as an obstacle. This criteria is working pretty well and a more thorough evaluation will be provided in the next section. In general those criteria help that the obstacles can be detected in any orientation. The only danger to the yellow detecting algorithm is motion blur, namely when the single lines are not separated but connected together by "blur".
\item After analyzing each of the potential obstacle objects, we decide whether it is an obstacle or not. If so, we continue to steps 7. and 8..
\item Afterwards, we calculate the position and radius of all of the obstacles. After segmenting the object we calculate the 4 corners (which are connected in to form the green rectangle). We defined the obstacle's position as the midpoint of the lower line (this point surely lies on the ground plane). For the radius, we use the distance in the real world between this point and the lower right corner. This turned out to be a good approximation of the radius. For an illustration you can have a look at figure~\ref{fig:position_size}.
\begin{figure}[ht]
  \includegraphics[width=1.0\linewidth]{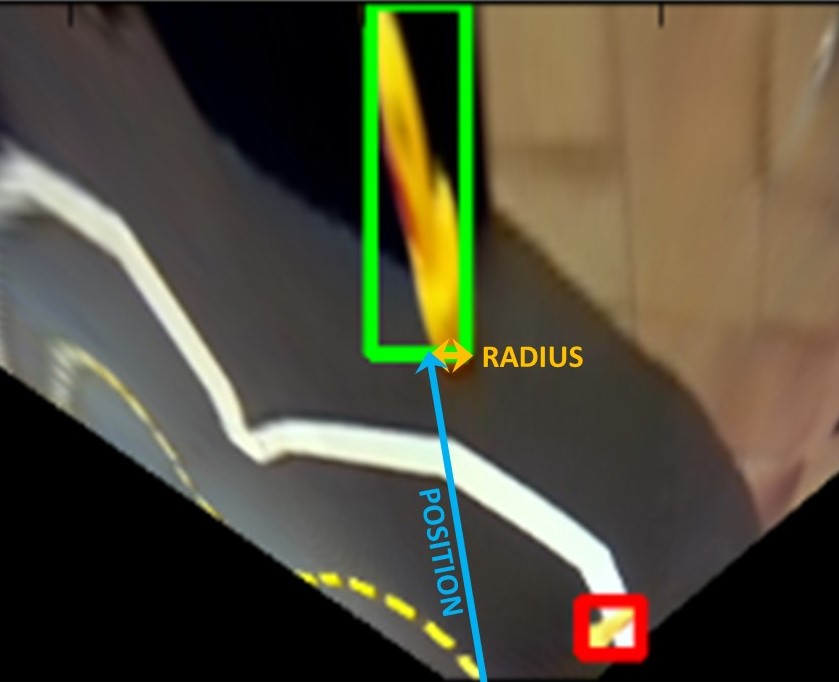}
  \caption{Position and Radius of the Obstacle}
  \label{fig:position_size}
\end{figure}
\item Towards the end of the project we came up with one additional last step based on the idea that only obstacles inside the white lane boundaries are of interest to us. That is why for each obstacle, we look whether there is something white in between us and the obstacle. In figure~\ref{fig:classification} you can see an example situation where the obstacle inside the lane is marked as dangerous (red) while the other one is marked as not of interest to us since it is outside the lane boundary (green). In figure~\ref{fig:search_line} you see the search lines (yellow) along which we search for white elements.
\begin{figure}[ht]
  \includegraphics[width=1.0\linewidth]{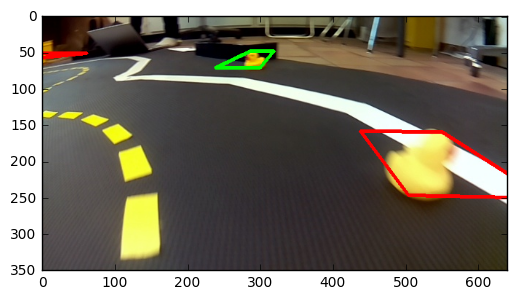}
  \caption{Classification if Objects are Dangerous or Not}
  \label{fig:classification}
\end{figure}
\begin{figure}[ht]
  \includegraphics[width=1.0\linewidth]{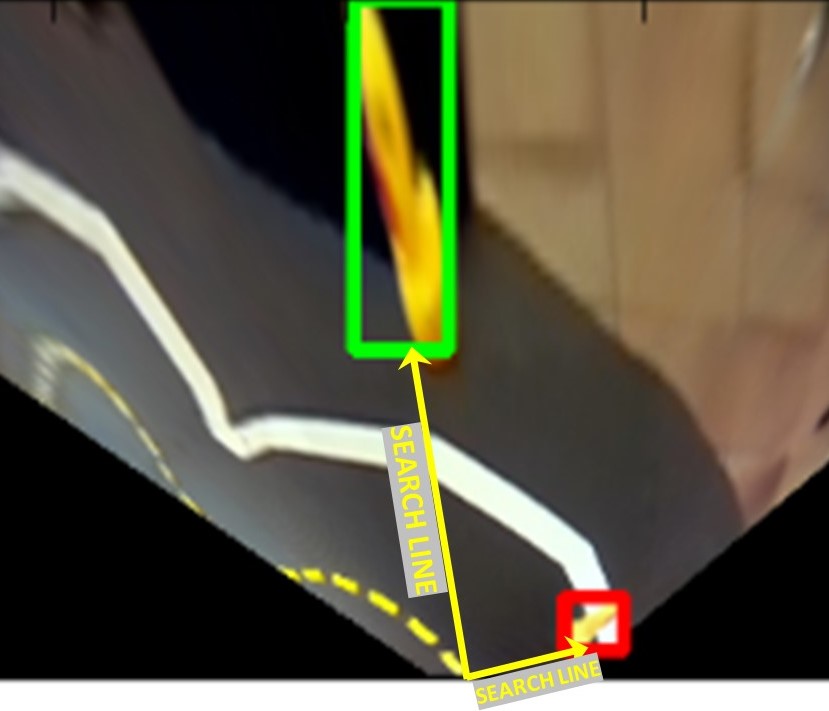}
  \caption{Search Lines to Infer if something White is in Between}
  \label{fig:search_line}
\end{figure}
\item As the last step of the detection pipeline we return a list of all obstacles including all the information via the \emph{posearray}.
\end{enumerate}

\section{Avoidance - Acting in the Real World}
\label{sec:avoidance}

\subsection{Description}

With the 3D position, size and the labeling whether the object is inside the lane boundaries or not given from section~\ref{sec:detection}, we wanted to reach the final objectives:
\begin{itemize}
\item Plan path around obstacle if possible (we have to stay within our lane).
\item If this is not possible, simply stop.
\end{itemize}

\subsubsection{Assumptions}

The assumptions for correctly reacting to the previously detected obstacles are:
\begin{itemize}
\item Heading and position relative to track given,
\item "The Controllers" are responsible for following our trajectory and 
\item we have the possibility to influence vehicle speed (slow down, stop).
\end{itemize}
As we now know, the first assumption is normally not fulfilled. We describe in the functionality section~\ref{sec:avoidFunctionality} section why this comes out to be a problem.

\subsubsection{Performance Metrics}

For measuring the performance we used:
\begin{itemize}
\item Avoid/hit ratio
\item performed during changing light conditions.
\end{itemize}

\subsection{Functionality}
\label{sec:avoidFunctionality}

The Avoidance deals with drawing the right conclusions from the received data and forwarding it.

\subsubsection{Theoretical Description}

With the separation of the detection, an important part of the avoidance node is the interaction with the other work packages. We determined the need of getting information about the remaining Duckietown besides the detected obstacles. The obstacles need to be in relation to the track, in order to assess whether we have to stop, can drive around obstacles or if it is even already out of track. Due to other teams already working on the orientation within Duckietown, we deemed it best to not implement any further detections (lines, intersections etc.) in our visual perception pipeline. This saves similar algorithms being run twice on the processor. We decided to acquire the values of our current pose relative to the side lane, which is determined by the \emph{line detection group}.

The idea was to make the system highly flexible. The option to adapt to following situations was deemed desirable:
\begin{itemize}
\item Multiple obstacles. Different path planning in case of a possible avoidance might be required.
\item Adapted behavior if the robot is at intersections.
\item Collision avoidance dependent on the fleet status within the Duckietown. Meaning, if a Duckiebot drives alone in a town it should have the option to avoid a collision by driving onto the opposite lane.
\end{itemize}
Obstacles sideways of the robot were expected to appear as the Duckietowns tend to be flooded by duckies. Those detections on the side as well as far away false positive detections should not make the robot stop. To prevent that, we intended on implementing a parametrized \textbf{bounding box} ahead of the robot. Only obstacles within that box would be considered. Depending on the certainty of the detections as well as the yaw-velocities the parametrization would be tuned.

The interface getting our computed desired values to impact the actual Duckiebot is handled by the \emph{controllers}. We agreed on the usage of their custom message format, in which we send desired values for the lateral lane position and the longitudinal velocity. Our intention was to \textbf{account for the delay} of the physical system in the \emph{avoider} node. Thus our planned trajectory will reach the offset earlier than the ideal-case trajectory would have to.

Due to above mentioned interfaces and multiple levels of goals, we were aiming for an architecture which allows \textbf{gradual commissioning}. The intent was to be able to go from basic to more advanced for us as well as for groups in upcoming years. Those should be able to extend our framework and not have to rebuild it.

\begin{figure}[ht]
  \includegraphics[width=1.0\linewidth]{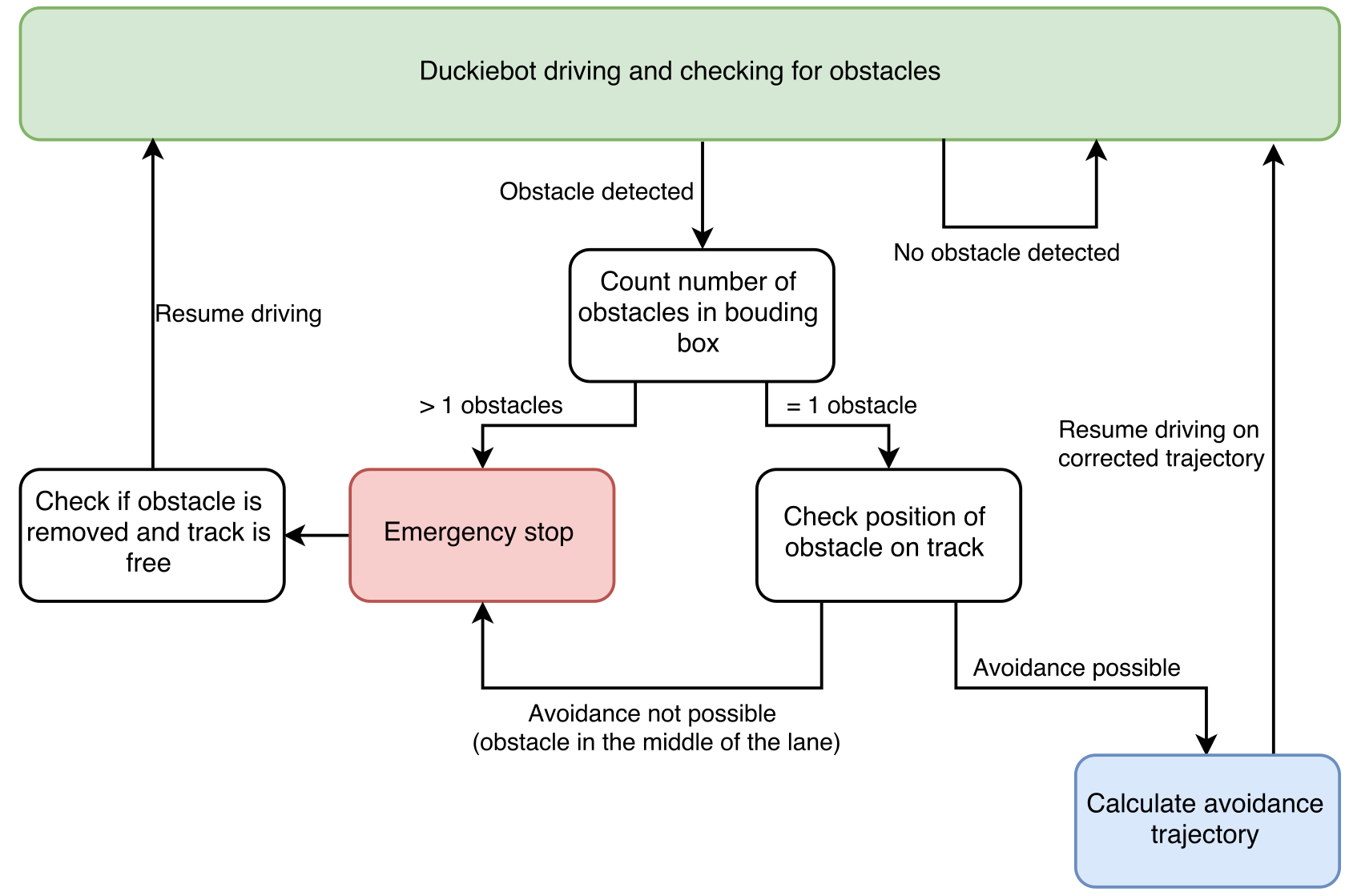}
  \caption{Logic of one of the First Stages in Commissioning}
  \label{fig:avoidance_logic}
\end{figure}

The logic shown in figure~\ref{fig:avoidance_logic} displays one of the first stages in the commissioning. Key is the reaction to the number of detected obstacles. Later stages will not trigger an emergency stop in case of multiple obstacle detections within the bounding box.

Our biggest concern were the added inaccuracies until the planning of the trajectory. Those include:
\begin{itemize}
\item Inaccuracy of the currently determined pose,
\item inaccuracy of the obstacle detection and
\item inaccuracy of the effectively driven path aka. controller performance.
\end{itemize}
To us, the determination of the pose was expected to be the most critical. Our preliminary results of the obstacle detection seemed reasonably accurate. The controller could be tweaked that the robot would rather drive out of the track than into the obstacle. Though, an inaccurate estimation of the pose would just widen the duckie artificially.

The \emph{controllers} did not plan on being able to intentionally leave the lane. Meaning the space left to avoid an obstacle on the side of the lane is tight making above uncertainties more severe.

We evaluated the option to keep track of our position inside the map. Given a decent accuracy of said position, we'd be able to create a map of the detected obstacles. Afterwards - especially given multiple detections (also outside of the bounding box) - we could achieve a further estimation of our pose relative to the obstacles. This essentially would mean creating a \emph{SLAM-algorithm} with obstacles as landmarks. We declared this as out of scope given the size of our team as well as the computational constraints. The goal was to make use of a stable, continuous detection and in each frame react on it.

\subsubsection{Actual Implementation}

\begin{figure}[ht]
  \includegraphics[width=1.0\linewidth]{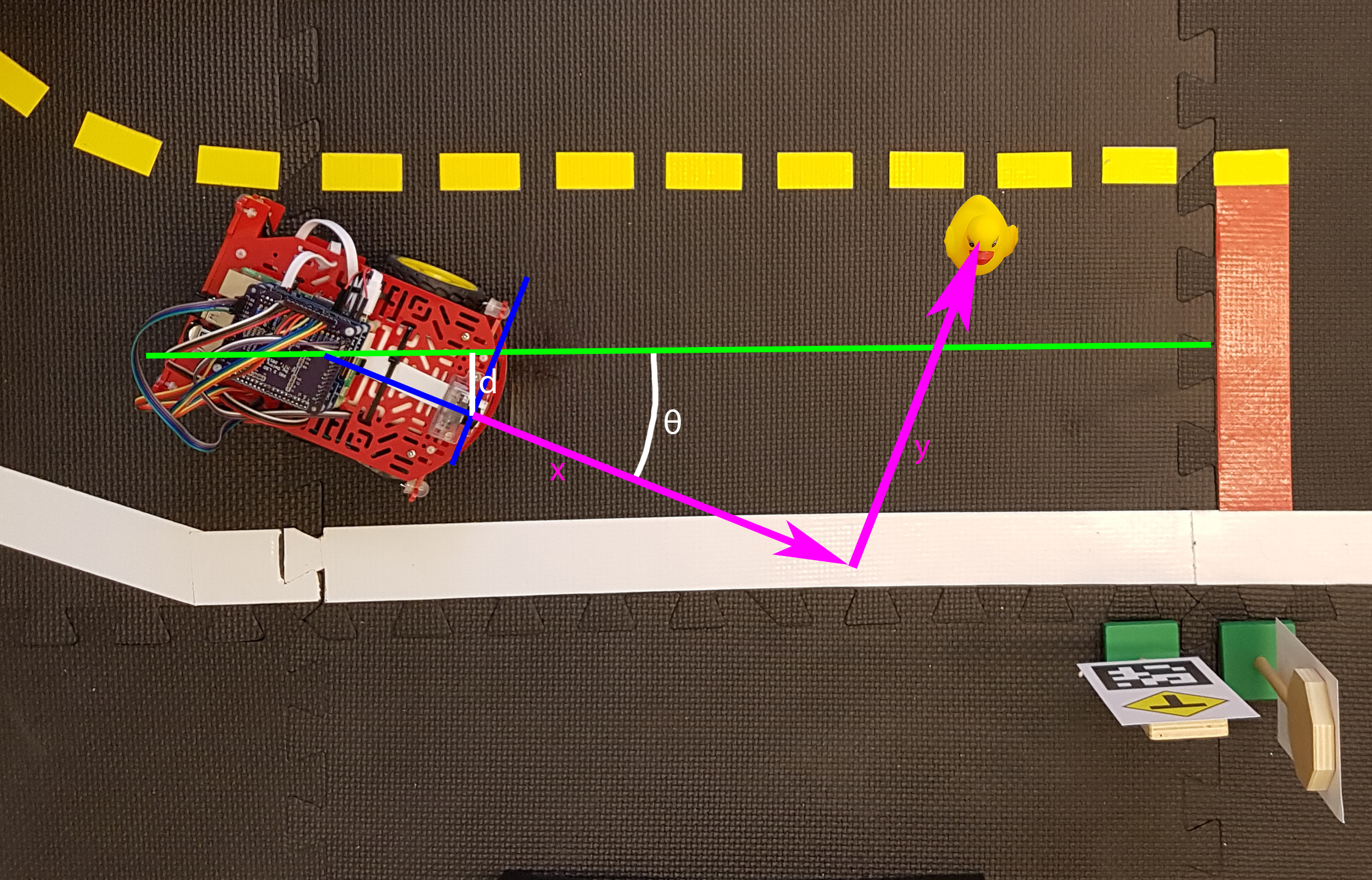}
  \caption{Variable Definitions seen from the Top}
  \label{fig:definitions_top}
\end{figure}

\textbf{Interfaces:} One important part of the Software is the handling of the interfaces, mainly to \emph{devel\_controllers}. For further information on this you can refer to the Software Architecture Chapter (section~\ref{sec:software_arch}). \newline
\textbf{Reaction:} The obstacle avoidance part of the problem is handled by an additional node, called the \emph{obstacle\_avoidance\_node}. The node uses two main inputs which are the \emph{obstacle pose} and the \emph{lane pose}. The obstacle pose is an input coming from the obstacle detection node, which contains an array of all the obstacles currently detected. Each array element consists of an x and y coordinate of an obstacle in the robot frame (with the camera as origin) and the radius of the detected object. By setting the radius to a negative value, the detection node indicates that this obstacle is outside the lane and should not be considered for avoidance. The lane pose is coming from the line detection node and contains among other unused channels the current estimated distance to the middle of the lane d as well as the current heading of the robot $\theta$. Figure~\ref{fig:definitions_top} introduces the orientations and definitions of the different inputs which are processed in the obstacle avoidance node.

Using the obstacle \emph{pose array} we determine how many obstacles need to be considered for avoidance. If the detected obstacle is outside the lane and therefore marked with a negative radius by the obstacle detection node we can ignore it. Furthermore, we use the before mentioned \textbf{bounding box} with tunable size which assures that only objects in a certain range from the robot are considered. As soon as an object within limits is inside of the bounding box, the \emph{obstacle\_avoidance\_active flag} is set to true and the algorithm already introduced in figure~\ref{fig:avoidance_logic} is executed.

\begin{figure}[ht]
  \includegraphics[width=1.0\linewidth]{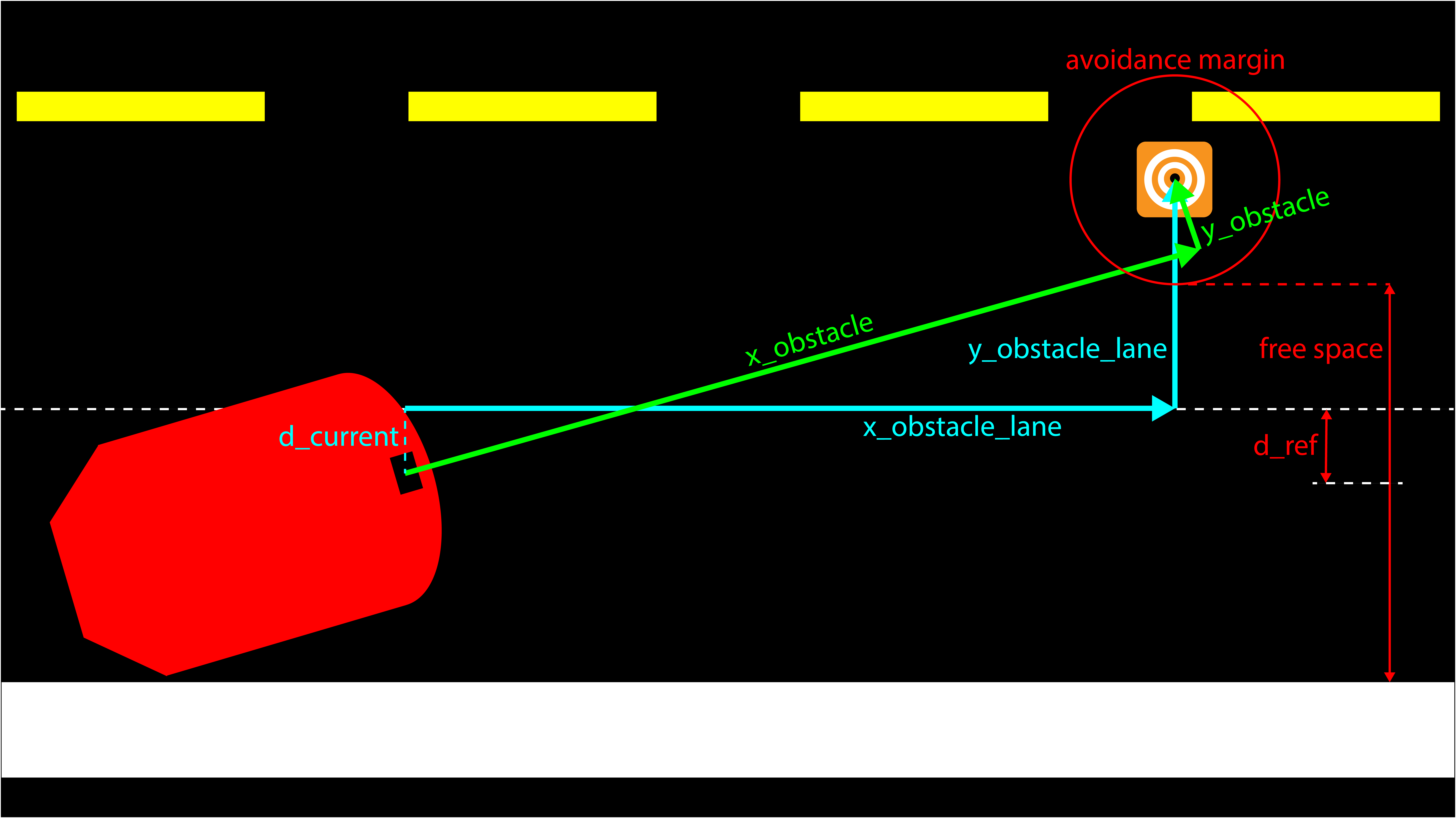}
  \caption{Geometry of described Scene}
  \label{fig:coordinate}
\end{figure}

\textbf{Case 1: Obstacle Avoidance} \newline
If there is only one obstacle in range and inside the bounding box, the obstacle avoidance code in the avoider function is executed. First step of the avoider function is to transform the transmitted obstacle coordinates from the robot frame to a frame which is fixed to the middle of the lane using the estimated measurements of $\theta$ and d. Doing this transformation allows us to calculate the distance of the object from the middle line. If the remaining space (in the lane (subtracted by a safety avoidance margin) is large enough for the robot to drive through, we proceed with the obstacle avoidance, if not we switch to case 2 and stop the vehicle. Please refer to figure\ref{fig:coordinate}. 
If the transformation shows that an avoidance is possible we calculate the d\_ref we need to achieve to avoid the obstacle. This is sent to the lane control node and then processed as new target distance to the middle of the lane. The lane control node uses this target and starts to correct the Duckiebot's position in the lane. With each new obstacle pose being generated this target is adapted so that the Duckiebot eventually reaches target position. The slow transition movement allows us to avoid the obstacle even when it is not visible anymore shortly before the robot is at the same level as the obstacle.

At the current stage, the obstacle avoidance is not working due to very \textbf{high inaccuracies} in the estimation of $\theta$. The value shows inaccuracies with an amplitude of 10$^{\circ}$, which leads to wrong calculations of the transformation and therefore to misjudgement of the d\_ref. The high amplitude of these imprecisions could be transformed to a uncertainty factor of around 3 which means that each object is around 3 times its actual size which means that even a small obstacle on the side of the lane would not allow a safe avoidance to take place. For this stage to work, the estimation of $\theta$ would need significant improvement.

\textbf{Case 2: Emergency Stop} \newline
Conditions for triggering an emergency stop:
\begin{itemize}
\item More than one obstacle in rang,
\item avoidance not possible because the obstacle is in the middle of the lane, 
\item currently every obstacle detection in the bounding box triggers an emergency stop due to the above mentioned uncertainty.
\end{itemize}
If one of the above scenarios occurs, an avoidance is not possible and the robot needs to be stopped. By setting the target speed to zero, the lane controller node stops the Duckiebot. As soon as the situation is resolved by removing the obstacle which triggered the emergency stop, the robot can proceed with the lane following. These tasks are then repeated at the frame rate of the obstacle detection array being sent.

\section{Software Architecture}
\label{sec:software_arch}

\begin{figure*}[t]
\centering
  \includegraphics[width=0.8\textwidth]{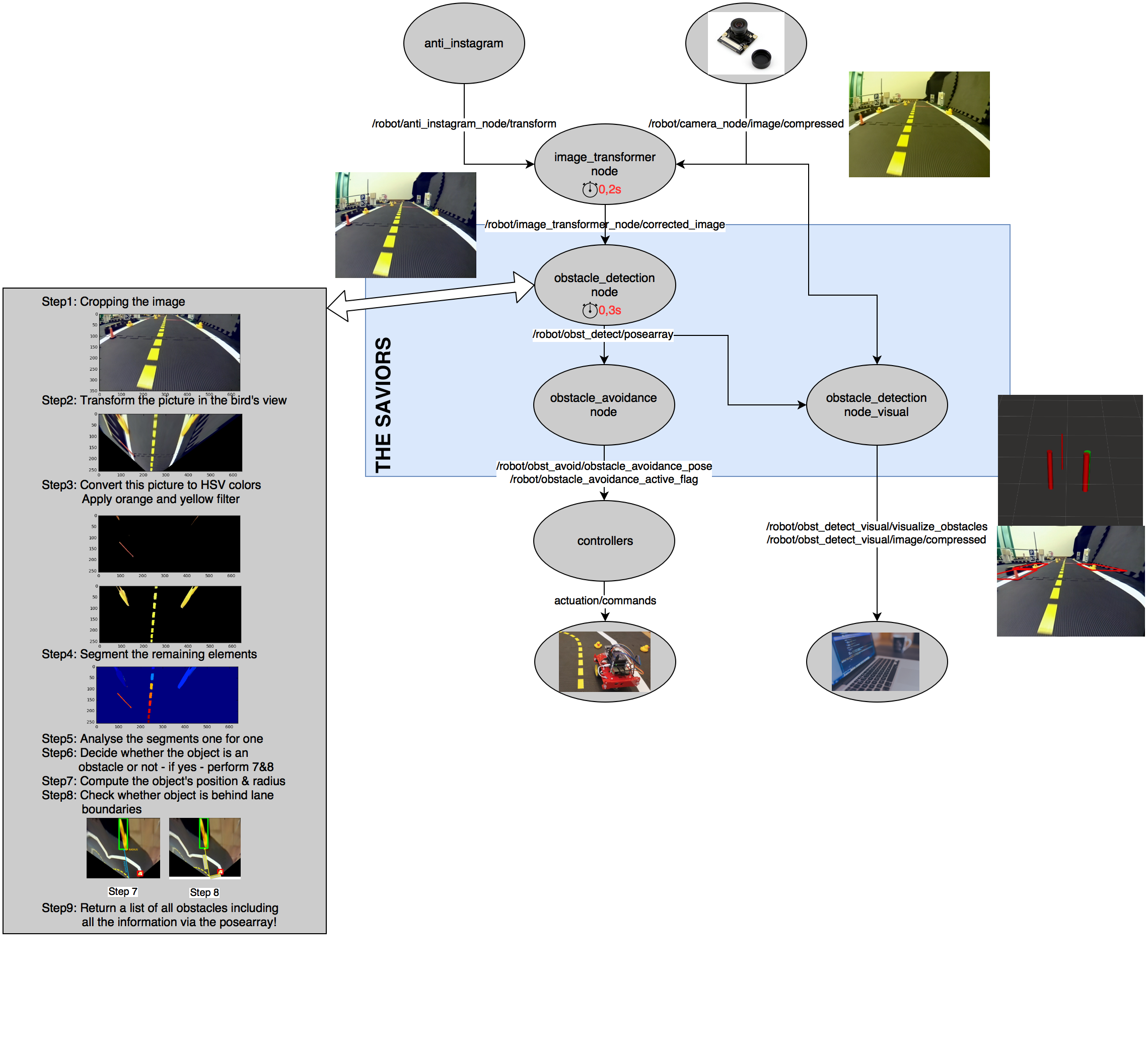}
  \caption{Overview of Software Architecture}
  \label{fig:overviewSoftware}
\end{figure*}

In general we have four interfaces which had to be created throughout the implementation of our software:
\begin{enumerate}
\item At first, we need to receive an incoming picture which we want to analyze. As our chosen approach includes filtering for specific colors, we are obviously dependent on the \textbf{lighting conditions}. In a first stage of our project, we nevertheless simply subscribed to the raw camera image because of the considerable expense of integrating the \emph{Anti Instagram Color Transformation} and since the Anti Instagram team also first had to further develop their algorithms. During our tests we quickly recognized that our color filtering based approach would always have some troubles if we don't compensate for the lighting change. Therefore, in the second part of the project we closely collaborated with the Anti Instagram team and are now subscribing to a color corrected image provided by them. Currently, to keep computational power on our Raspberry Pi low, the corrected image is published at 4 Hz only and the color transformation needs at most 0.2 seconds.
\item The second part of our System Integration is the \textbf{internal interface} between the object detection and avoidance part. The interface is defined as a \emph{PoseArray} which has the same time stamp as the picture from which the obstacles have been extracted. This Array, as the name already describes, is made up of single poses. The meaning of those are the following:

The position x and y describe the real world position of the obstacle which is in our case the center front coordinate of the obstacle. Since we assume planarity, the z coordinate of the position is not needed. That is why we are using this z coordinate to describe the radius of the obstacle.

Furthermore a negative z coordinate shows that there is a white line in between us and the obstacle which indicates that it is not dangerous to us since we assume to always having to stay in the lane boundaries. Therefore this information allows us to not stop if there is an obstacle behind a turn.

As for the scope of our project, the orientation of the obstacles is not really important, we use the remaining four elements of the Pose Message to pass the pixel coordinates of the bounding box of the obstacle seen in the bird view. This is not needed for our "reaction" module but allows us to implement an efficient way of visualization which will be later described in detail. Furthermore, we expect our obstacle detection module to add an additional delay of about max. 0.3s.
\item The third part is the interface between our obstacle avoidance node and the \textbf{Controllers}. The obstacle avoidance node generates an obstacle avoidance \emph{PoseArray} and obstacle avoidance active flag.

The obstacle avoidance \emph{PoseArray} is the main interface between the Saviors and the group doing lane control. We use the \emph{PoseArray} to transmit d\_ref (target distance to middle of the lane) and v\_ref (target robot speed). The d\_ref is our main control output which enables us to position the robot inside the lane and therefore to avoid objects which are placed close the lane line on the track. Furthermore v\_ref is used to stop the robot when there is an unavoidable object by setting the target speed to zero.

The flag is used to communicate to the lane control nodes when the obstacle avoidance is activated which then triggers d\_ref and v\_ref tracking.
\item The fourth part is an optional interface between the Duckiebot and the user's \textbf{personal laptop}. Especially for the needs of debugging and inferring what is going on, we decided to implement a visualization node which can visualize on the one hand the input image including bounding boxes around all the objects which were classified as obstacles and furthermore this node can output the obstacles as markers which can be displayed in rviz.
\end{enumerate}
In figure~\ref{fig:overviewSoftware} you find a graph which summarizes our software packages and gives a brief overview.

\section{Visualizer}

Especially when dealing with a visual based obstacle detection algorithm it is very hard to infer what is going on. One has to also keep the visual outputs low, to consume as less computing power as possible, especially on the Raspberry Pi. This is why we decided to not implement one single obstacle detection node, but effectively two of them, together with some scripts which should help to tune the parameters offline and to infer the number of false positives, etc.. The node which is designed to be run on the Raspberry Pi is our normal obstacle\_detection\_node. This should in general be run such that there is no visual output at all but that simply the \emph{PoseArray} of obstacles is published through this node.

The other node, namely the \emph{obstacle\_detection\_visual\_node} is designed to be run on your own laptop which is basically visualizing the information given by the \emph{PoseArray}. There are two visualisations available. On the one hand there is a marker visualization in rviz which shows the position and size of the obstacles. In here all the dangerous obstacles which must be considered are shown in red, whereas the non critical (which we think that they are outside the lane boundaries) are marked in green. On the other hand there is also a visualization available which shows the camera image together with bounding boxes around the detected obstacles. Nevertheless, this online visualization is still dependent on the connectivity and you can only hardly "freeze" single situations where our algorithm failed. That is why we also included some helpful scripts into our package. One script allows to thoroughly input many pictures and outputs them labeled together with the bounding boxes, while another one outputs all the intermediate steps of our filtering process which allows to fastly adapt e.g. the color thresholds which is in our opinion still the major reason for failure. More information on our created scripts can be found in our Readme on GitHub \footnote{\url{https://github.com/duckietown/Software/blob/devel-saviors-23feb/catkin_ws/src/25-devel-saviors/obst_avoid/README.md}}.

\section{Performance Evaluation}
\label{sec:perfEval}

\subsection{Evaluation of the Interface and Computational Load}

In general, as we are dealing with many color filters a reasonable color corrected image is the key to the good functioning of our whole module, but turned out to be the greatest challenge when it comes down to computational efficiency and performance. As described above we are really dependent on a color corrected image by the Anti Instagram module. Throughout the whole project we planned to use their continuous anti-instagram node which is supposed to compute a color transformation in fixed intervals of time. However, when it came down we actually had to change this for the following reason: The continuous anti-instagram node, running at an update interval of 10 seconds, consumes a considerable amount of computing power, namely 80\%. In addition to that, the image transformer node which is in fact transforming the whole image and currently running at 4 Hz needs another 74\% of one kernel. If you now run those two algorithms combined with the lane-following demo which makes the vehicle move and combined with our own code which needs an additional 75\% of computing power, our safety critical module could only run at 1.5Hz and resulted in poor behavior.

Even if you increase the time interval in which the continuous anti-instagram node computes a new transformation there was no real improvement. That is why in our final setup we let the anti-instagram node once compute a reasonable transformation and then keep this one for the entire drive. Through this measure we were able to safe the 80\% share entirely and this allowed our overall node to be run at about 3 Hz with introducing an additional maximal delay of about 0.3 seconds. Nevertheless we want to point out that all the infrastructure for using the continuous anti instagram node in the future is provided in our package.

To sum up, the interface between our node and the Anti Instagram node was for sure developed very well and the collaboration was very good but when it came to getting the code to work, we had to take one step back to achieve good performance. That is why it might be reasonable to put effort into this interface in the future, to being able to more efficiently transform an entire image and to reduce the computational power consumed by the node which continuously computes a new transformation.

\subsection{Evaluation of the Obstacle Detection}

In general, since our obstacle classification algorithm is based on the rotational invariant feature of the eigenvalues of the inertia tensor it is completely invariant to the current orientation of the duckiebot and its position with respect to the lanes.

To rigorously evaluate our detection algorithm, we started off with evaluating \textbf{static scenes}, meaning the Duckiebot is standing still and not moving at all. Our algorithm performed extremely well in those static situations. You can place an arbitrary amount of obstacles, where the orientation of the respective obstacles does not matter at all, in front of the Duckiebot. In those situations and also combining them with changing the relative orientation of the Duckiebot itself, we achieved a false positive percentage of below 1\% and we labeled all of the obstacles with respect to the lane boundaries correctly. The only static setup which is sometimes problematic is when we place the smallest duckies very close in front of our vehicle (below 4 centimeters), without approaching them. Then we sometimes cannot detect them. However this problem is mostly avoided during the dynamic driving, since we anyways want to stop earlier than 4 centimeters in front of potential obstacles. We are very happy with this static behavior as in the worst case, if during the dynamic drive something goes wrong, you can still simply stop and rely upon the fact that the static performance is very good before continuing your drive. In the log chapter it is possible to find the corresponding logs.

\begin{figure}[ht]
  \includegraphics[width=1.0\linewidth]{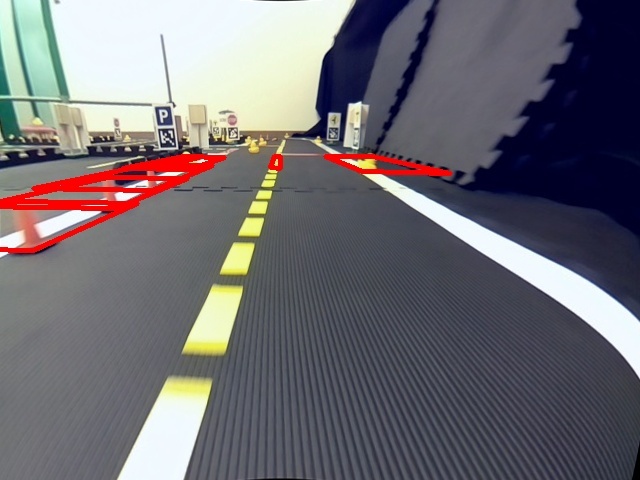}
  \caption{Obstacle Detector Error due to Motion Blur}
  \label{fig:motion_blur_error}
\end{figure}

This in return also implies that most of the misclassification errors during our \textbf{dynamic drive} are due to the effect of motion blur, assuming a stable color transformation provided by the anti instagram module. E.g. in figure~\ref{fig:motion_blur_error}, two line segments in the background "blurred" together for two consecutive frames resulting in being labeled as an obstacle.

Speaking more about of numbers, we took 2 duckiebots at a gain of around 0.6 and performed two drives at different days, so also at different lights and the results are the following: Evaluating each picture which will be given to the algorithm, we found out that on average, we detect 97\% of all the yellow duckies in each picture. In terms of cones we detect about 96\% of all cones in the evaluated frames. We consider these to be very good results as we have a very low rate of false positives (below 3\%). For further information look at table~\ref{table:results}.
\begin{table*}[ht]
  \small
  \centering
  \begin{tabular}{|c|c|c|c|c|c|c|}
    \hline
    \textbf{Date} & \textbf{correctly detected duckies} & \textbf{correctly detected cones} & \textbf{missed ducks} & \textbf{missed cones} & \textbf{false positive} & \textbf{false position} \\ \hline
    19.12.2017 & 423 & 192 & 14    & 8     & 9      & 45    \\
               &     &     & 3.2\% & 4\%   & 1.4\%  & 7.2\% \\ \hline
    21.12.2017 & 387 & 103 & 10    & 5     & 15     & 28    \\
               &     &     & 2.5\% & 4.4\% & 3\%    & 5.7\% \\ \hline  
  \end{tabular}
  \caption{Results}
  \label{table:results}
\end{table*}
When it comes to evaluating the performance of our obstacle classification with respect to classifying them as dangerous or not dangerous our performance is not as good as the detection itself, but we did also not put the same effort into it. As you can see in table~\ref{table:results}, we have an error rate of above 5\% when it comes to determining whether the obstacle's position is inside or outside the lane boundaries (this is denoted as false position in the table above). We are especially encountering problems when there is direct illumination on the yellow lines which are very reflective and therefore appear whitish. shows such a situation where the current implementation of our obstacle classification algorithm fails.

\subsection{Evaluation of the Obstacle Avoidance}

Since at the current state we stop for every obstacle which is inside the lane and inside the bounding box, the avoidance process is very stable since it does not have to generate avoidance trajectories. The final performance on the avoidance is mainly relying on the placement of the obstacles:
\begin{enumerate}
\item \textbf{Obstacle placement on a straight:} If the obstacle is placed on a straight with a sufficient distance from the corner the emergency stop works nearly every time if the obstacle is detected correctly.
\item \textbf{Obstacle in a corner:} Due to the currently missing information of the curvature of the current tile the bounding box is always rectangular in front of the robot. This leads to problems if an obstacle is placed in a corner because it might enter the bounding box very late (if at all). Since the detection very close to the robot is not possible, this can lead to crashes.
\item \textbf{Obstacles on intersection:} These were not yet considered in our scope but still work if the detection is correct. It then behaves similar to case 1.
\end{enumerate}

Furthermore there a few cases which can lead to problems independent of the obstacle placement: 
\begin{enumerate}
\item \textbf{Controller oscillations:} If the lane controller sees a lot of lag due to high computing loads or similar its control sometimes start to oscillate. These oscillations lead to a lot of motion blur which can induce problems in the detection and shorten the available reaction time to trigger an emergency stop.
\item \textbf{Controller offset:} The current size of the bounding box assumes that the robot is driving in the middle of the lane. If the robot is driving with an offset to the middle of the lane it can happen that obstacles at the side of the lane aren't detected. This however rarely leads to crashes because then the robot is just avoiding the obstacle instead of stopping for it.
\end{enumerate}

While testing our algorithms we saw successfull emergency stops in 10/10 cases for obstacles on a straight and in 3/10 cases for obstacles placed in corners assuming that the controller was acting normally. It is to be noted that the focus was lying on the reliable detections on the straights, which we intended to show on the demo day.

\section{Overall Result}

The final result can be seen in this video: \url{https://player.vimeo.com/video/251523150}.

\section{Outlook}

As already described above in the eval interface section, we think that there is still room for improving the interface between our code and the Anti Instagram module in terms of making the continouus anti instagram node as well as the image transformer node more computationally efficient. Another interesting thought which might be taken into consideration concerning this interface is the following: As long as the main part of the anti instagram's color correction is linear (which was in most of our cases sufficient), it might be reasonable to just adapt the filter values than to subscribe to a fully transformed image. This effort could save a whole publisher and subscriber and it is obvious that it is by far more efficient to transform a few filter values once than to transform every pixel of every incoming picture. Towards the end of our project we invested some time in trying to get this approach to work but as time was not enough we could not make it. We especially struggled to transform the orange filter values, while it worked for the yellow ones (BRANCH: devel-saviors-ai-tryout2). We think that if in the future one will stick to the current hardware this might be a very interesting approach, also for other software components such as the lane detection or any other picture related algorithms which are based on the concept of filtering colors.

Another idea of our team would be to \textbf{exploit the transformation to the bird's view also for other modules}. We think that this approach might be of interest e.g. for extracting the curvature of the road or performing the lane detection from the rather more undistorted top view.

Another area of improvement would be to further develop our provided scripts to being able to \textbf{automatically evaluate} the performance of our entire pipeline. As you can see in our code description in github there is a complete set of scripts available which makes it easily possible to transform a bag of raw camera images to a set of pictures on which we applied our obstacle detector, including the color correction part of Anti Instagram. The only missing step left is an automatic detection whether the drawn box is correct and in fact around an object which is considered to be an obstacle or not.

Furthermore to achieve more general performance propably even adaptions in the hardware might be considered to tune the obstacle detection algorithm and especially its generality. We think that setting up a \textbf{neural network} might make it possible to release the restrictions on the color of the obstacles (see \cite{2017} for further information).

In terms of avoidance there would be \textbf{possibilities to handle the high inaccuracies of the pose estimation} by relying on the lane controller to not leave the lane and just use a kind of closed loop control to avoid the obstacle (use the new position of the detected obstacle in each frame to securely avoid the duckie). Applying filters to the signals, especially the heading estimation, could further improve the behavior. This problem was detected late in the development and could not be tested due to time constraints. Going further, having both the line and obstacle detection in the same algorithm would allow the direct information on how far away obstacles are from the track. We expect that this would increase the accuracy compared to computing each individually and bringing it together.

The infrastructure is in place to include new scenarios like obstacles on intersection or multiple detected obstacles inside the bounding box. If multiple obstacles are in proximity, a more sophisticated trajectory generation could be put in place to avoid these obstacles in a safe and optimal way.

\begin{figure}[ht]
  \includegraphics[width=1.0\linewidth]{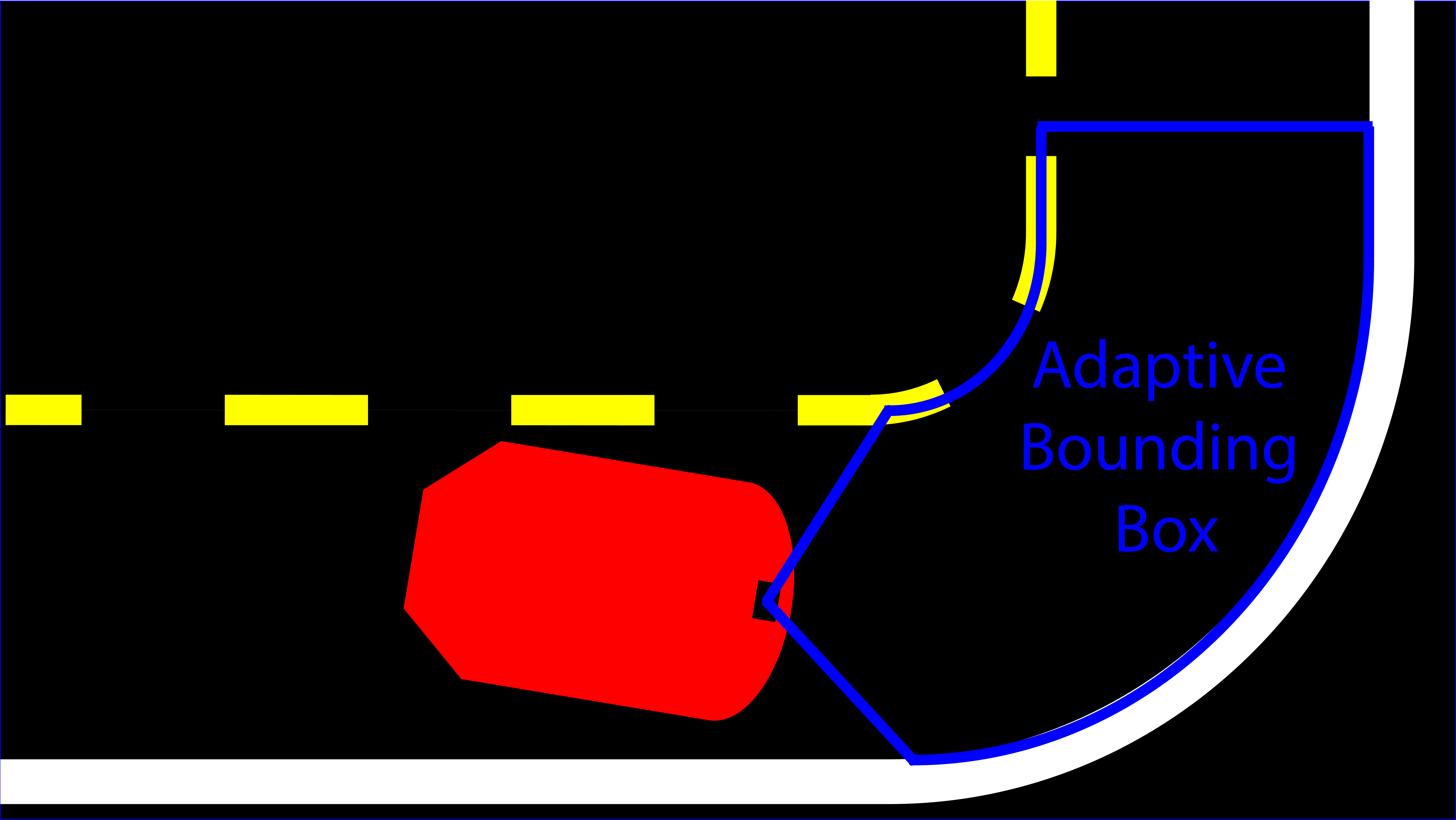}
  \caption{Adaptive Bounding Box}
  \label{fig:adaptive_bbox}
\end{figure}

Furthermore the \textbf{avoidance in corners} could be easily significantly improved if the line detection would estimate the curvature of the current tile which would enable adaptions to the bounding box on corner tiles. If the pose estimation is significantly improved one could also implement an adaptive bounding box which takes exactly the form of the lane in front of the robot (see figure~\ref{fig:adaptive_bbox}). 

\section{Theory}
\label{sec:theory}

\subsection{Computer Vision}

\subsubsection{Introduction}

\begin{figure}[ht]
  \includegraphics[width=1.0\linewidth]{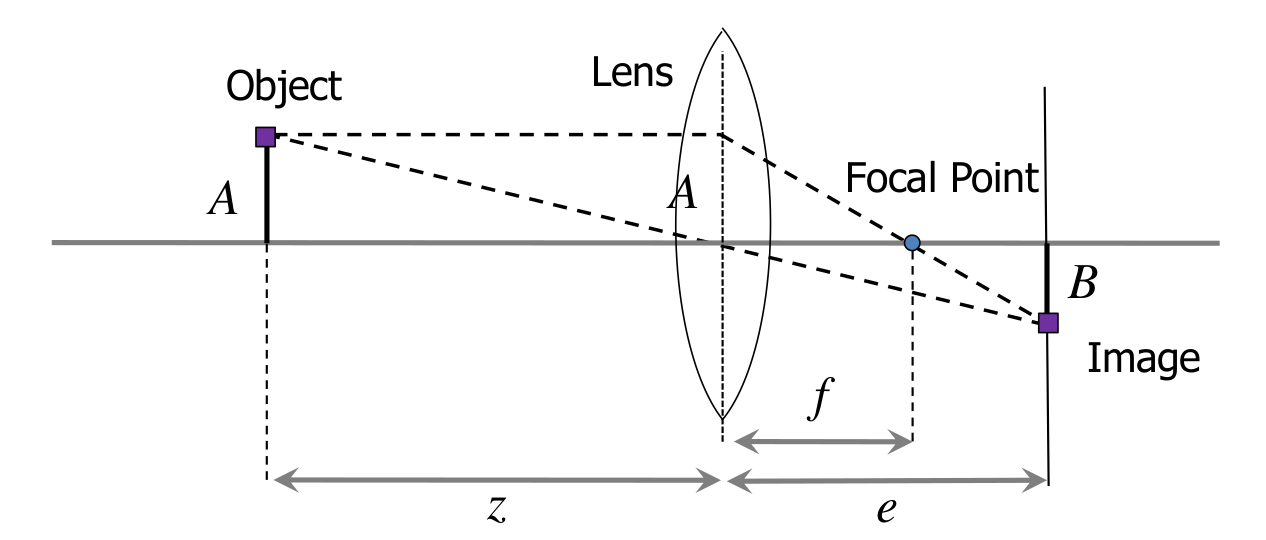}
  \caption{Simplified Camera Model \cite{Scaramuzz2017}}
  \label{fig:thin_lens}
\end{figure}

In general, a camera is consisting of a converging lens and an image plane (figure~\ref{fig:thin_lens}). In the following theory chapter, we will assume that the picture in the image plane is already undistorted, meaning we preprocessed it and eliminated the lens distortion.

\begin{figure}[ht]
  \includegraphics[width=1.0\linewidth]{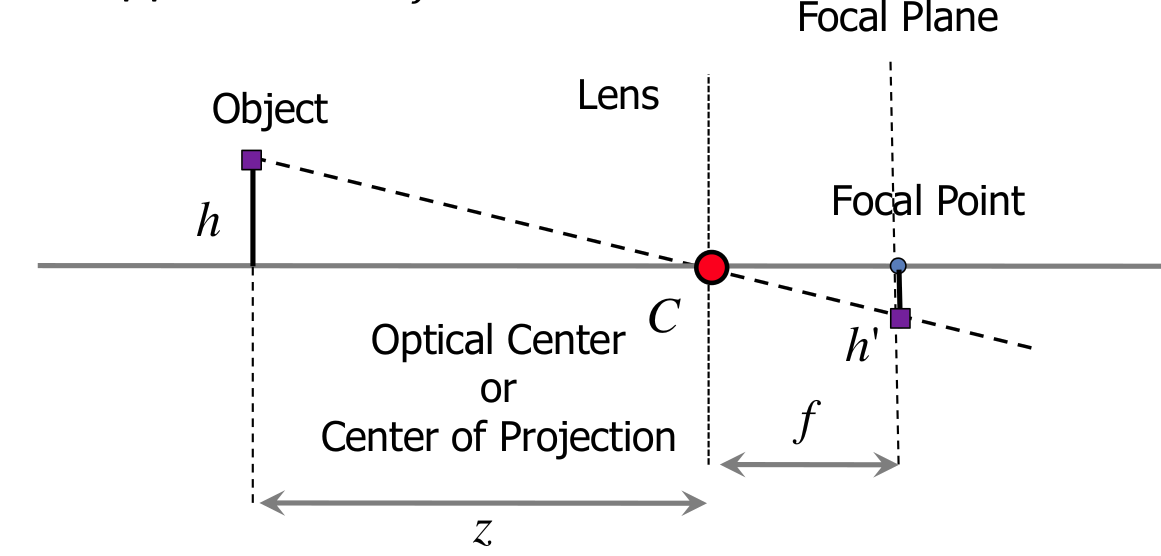}
  \caption{Pinhole Camera Approximation \cite{Scaramuzz2017}}
  \label{fig:pinhole_approx}
\end{figure}

It is quite easy to infer from figure~\ref{fig:thin_lens} that for a real world point to be in focus, it has to hold, that both of the "rays" intersect in one point in the image plane, namely in point B. Mathematically written this means:
\begin{equation}
\frac{B}{A}=\frac{e}{z} \quad and \quad \frac{B}{A}=\frac{e-f}{f}
\end{equation}\\
\begin{equation}
    \Leftrightarrow \frac{e}{f}-1=\frac{e}{z} \label{eq:two}
\end{equation}
This last equation~\ref{eq:two} can be approximated since usually $z \gg f$ such that we effectively arrive at the pin-hole approximation with: $e \approx f$ (see figure~\ref{fig:pinhole_approx}).

For the pixel coordinate on the image plane it holds:
\begin{equation}
\frac{h'}{h}=\frac{f}{z} \Leftrightarrow h'=\frac{f}{z} h
\end{equation}
In a more general case, when you consider a 3 dimensional setup and think of a 2 dimensional image plane you have to add another dimension and it follows that a real world point being at $ \vec{P_W} = (X_W, Y_W, Z_W)^T$ will therefore be projected to the pixels in the image plane:
\begin{equation}
x_{pix}=\frac{\alpha f}{Z_W} X_W + x_{offset}, y_{pix}=\frac{\beta f}{Z_W} Y_W + y_{offset}
\end{equation}
where $\alpha$ and $\beta$ are scaling parameters and $x_{offset}$ and $y_{offset}$ are constants which can be always added. Those equations are usually rewritten in homogeneous coordinates such that we have only linear operations left as:
\begin{equation} 
\Leftrightarrow \lambda \vec{P_{pix}} = H \vec{P_W} \label{eq:one} 
\end{equation}
\textbf{Note:} In general this Matrix H is what we get out of the intrinsic calibration procedure.

\begin{figure}[ht]
  \includegraphics[width=1.0\linewidth]{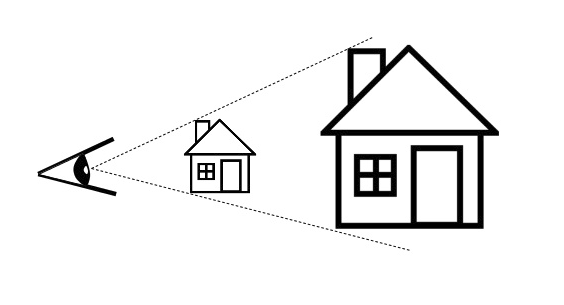}
  \caption{Scale Ambiguity \cite{Scaramuzz2015}}
  \label{fig:scale_amb}
\end{figure}

This equation~\ref{eq:one} and especially figure~\ref{fig:pinhole_approx} clearly show that since in every situation you only know H as well as $x_{pix}$ and $y_{pix}$ of the respective objects on the image plane, there is no way to determine the real position of the object, since everything can only be determined up to a scale ($\lambda$). Frankly speaking you only know the direction in which the object has to be but nothing more, which makes it a very difficult task to infer potential obstacles given the picture of a monocular camera only. This scale ambiguity is also illustrated in figure~\ref{fig:scale_amb}.

To conclude, given a picture from a monocular camera only, you have no idea at which position the house really is, so without exploiting any further knowledge it is extremely difficult to reliably detect obstacles which is also the main reason why the old approach did not really work. On top of that come other artifacts such as that the same object will appear larger if it is closer to your camera and vice versa, and lines which are parallel in the real world will in general not be parallel in your camera image.

\begin{figure}[ht]
  \includegraphics[width=1.0\linewidth]{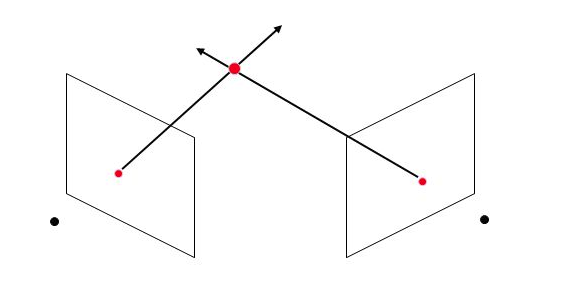}
  \caption{Triangulation to Obtain Absolute Scale \cite{Rasmussen}}
  \label{fig:triang}
\end{figure}

Note: The intuition, why we humans can infer the real scale of objects is that if you add a second camera, know the relative Transformation between the two cameras and see the same object in both images, then you can easily triangulate the full position of the object, since it is at the place where the two "rays" intersect (see figure~\ref{fig:triang}).

\subsubsection{Inverse Perspective Mapping / Bird's View Perspective}

\begin{figure}[ht]
\centering
  \includegraphics[width=0.8\linewidth]{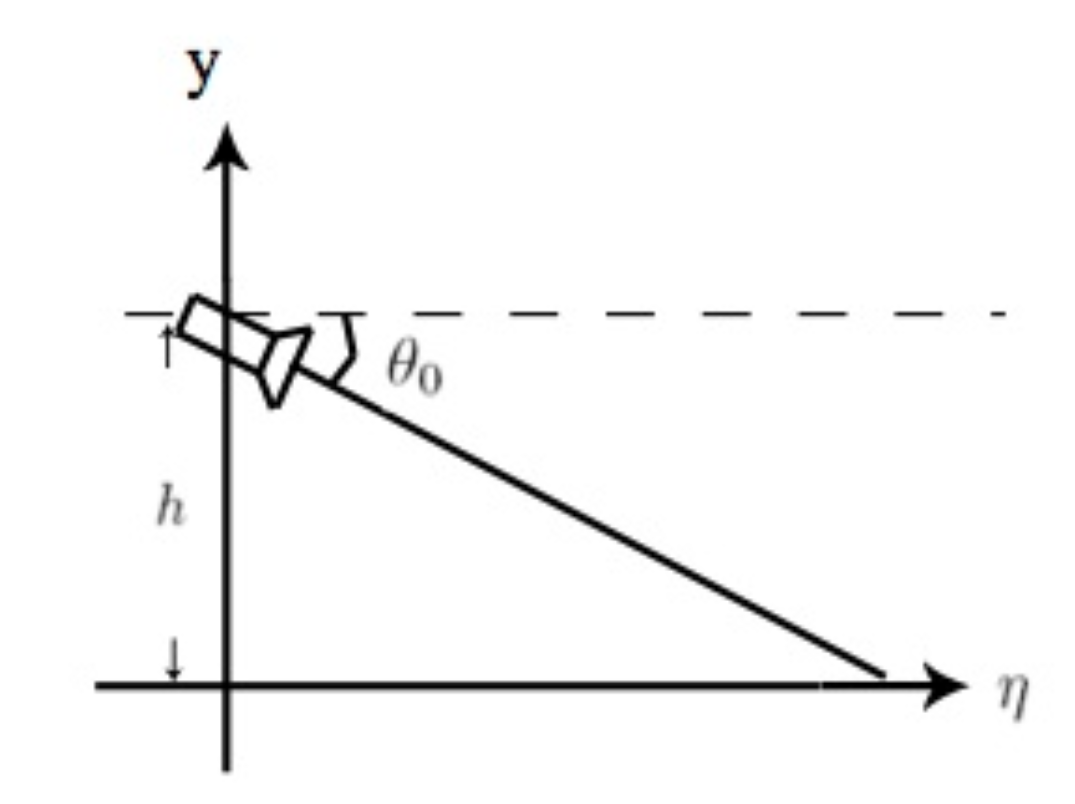}
  \caption{Illustration of our Fixed Camera Position \cite{2012}}
  \label{fig:fixed_cam}
\end{figure}

The first chapter above introduced the rough theory which is needed for understanding the follwing parts. The important additional information that we exploited heavily in our approach is that in our special case we know the coordinate $Z_W$. The reason therefore lies within the fact that unlike in another more general use case of a mono camera, we know that our camera will always be at height $h$ with respect to the street plane and that the angle $\theta_0$ also always stays constant (figure~\ref{fig:fixed_cam}).

This information is used in the actual extrinsic calibration (e.g. in Duckietown) due to the assumption that everything we see should in general be on the road. Therefore we can determine the full real world coordinates of every pixel, since we know the coordinate $Z_W$ which uniquely defines the absolute scale and can therefore uniquely determine $\lambda$ and H. Intuitively this comes from the fact that we can just intersect the known ray direction with the known "ground plane".

This makes it possible to project every pixel back into the "ground plane" by computing for each available pixel: 
\begin{equation}
\vec{P_W}=H^{-1} \lambda P_{pix}
\end{equation}

\textbf{This "projection back onto the ground plane" is called inverse perspective mapping!}

If you now visualize this "back projection", you basically get the bird's view since you can now map back every pixel in the image plane to a unique place on the road plane.

\begin{figure}[ht]
  \includegraphics[width=1.0\linewidth]{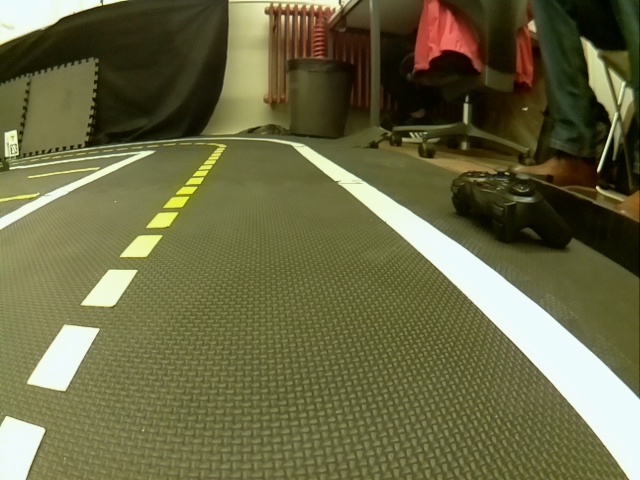}
  \caption{Normal incoming Image without any Obstacle}
  \label{fig:raw_img}
\end{figure}

\begin{figure}[ht]
  \includegraphics[width=1.0\linewidth]{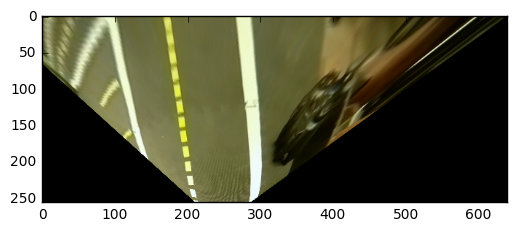}
  \caption{Incoming Image without any Obstacle Reconstructed in Bird's View}
  \label{fig:normal_bird}
\end{figure}

The only trick of this easy maths is that we exploited the knowledge that everything we see in the image plane is in fact on the road and has one and the same z-coordinate. You can see that the original input image (figure~\ref{fig:raw_img}) is nicely transformed into the view from above where every texture and shape is nicely reconstructed if this assumption is valid (figure~\ref{fig:normal_bird}). You can especially see that all the yellow line segments in the middle of the road roughly have the same size in this bird's view which is very different if you compare it to the original image.

\begin{figure}[ht]
\centering
  \includegraphics[width=0.8\linewidth]{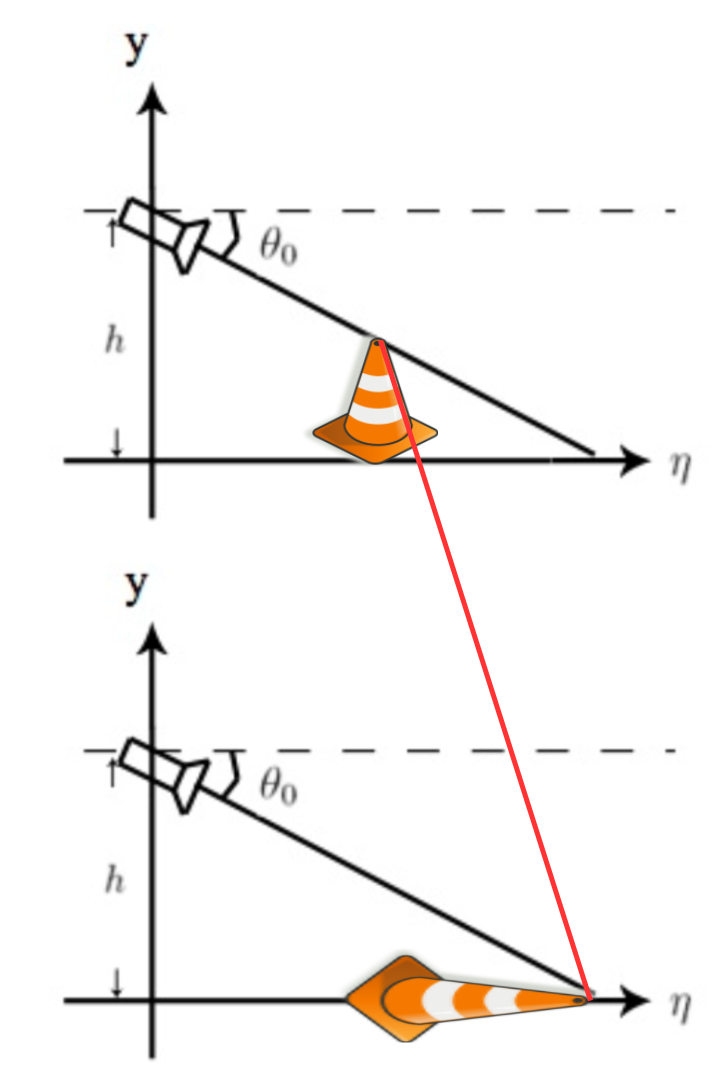}
  \caption{Illustration why Obstacle standing out of Ground Plane is heavily Disturbed in Bird's View (modified \cite{2012})}
  \label{fig:reason_for_dist}
\end{figure}

The crucial part is now what happens in this bird's view perspective, if the camera sees an object which is not entirely part of the ground plane, but stands out. These are basically obstacles we want to detect. If we still transform the whole image to the bird's view, these obstacles which stand out of the image plane get heavily distorted. Lets explain this by having a look at figure~\ref{fig:reason_for_dist}.

The upper picture depicts the real world situation, where the cone is standing out ot the image plane and therefore the tip is obviously not at the same height as the ground plane. However, as we still have this assumption and as stated above intuitively intersect the ray with the ground plane, the cone gets heavily disturbed and will look like the lower picture after performing the inverse perspective mapping. From this follows that if there are any objects which do stand out of the image plane then in the inverse perspective you basically see their shape being projected onto the ground plane. This behavior can be easily exploited since all of these objects are heavily distorted, drastically increase in size and can therefore be easily separated from the other objects which belong to the ground plane.

\begin{figure}[ht]
  \includegraphics[width=1.0\linewidth]{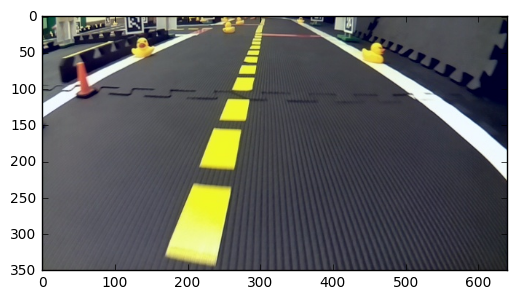}
  \caption{Normal Situation with Obstacles in Duckietown seen from Duckiebot Perspective}
  \label{fig:incoming_img}
\end{figure}

\begin{figure}[ht]
  \includegraphics[width=1.0\linewidth]{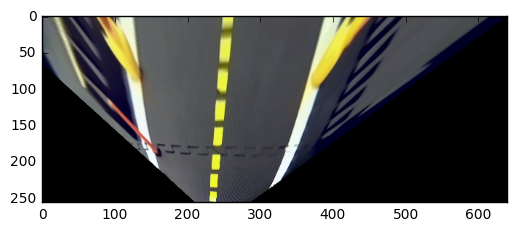}
  \caption{Same Situation seen from Bird's Perspective}
  \label{fig:bird_img}
\end{figure}

Let's have one final look at an example in Duckietown. In figure~\ref{fig:incoming_img} you see an incoming picture seen from the normal camera perspective, including obstacles. If you now perform the inverse perspective mapping, the picture looks like figure~\ref{fig:bird_img} and as you can easily see, all the obstacles, namely the two yellow duckies and the orange cone which stand out of the ground plane are heavily disturbed and therefore it is quite easy to detect them as real obstacles.

\subsection{HSV Color Space}

\subsubsection{Introduction and Motivation}

The "typical" color model is called the \emph{RGB} color model. It simply uses three numbers for the amount of the colors red, blue and green. It is an additive color system, so we can simply add two colors to produce a third one. Mathematically written it looks as follows and shows the way of how we deal with producing new colors:
\begin{equation}
(r_{res}, g_{res}, b_{res})^T = (r_{1}, g_{1}, b_{1})^T + (r_{2}, g_{2}, b_{2})^T
\end{equation}
If the resulting color is white, the two colors 1 and 2 are called to be complementary (e.g. this is the case for blue and yellow). This color system is very intuitive and is oriented on how the human vision perceives the different colors.

The \emph{HSV} color space is an alternative representation of the \emph{RGB} color model. On this occasion, \emph{HSV} is an acronym for \emph{Hue}, \emph{Saturation} and \emph{Value}. It is not so easy summable as the \emph{RGB} model and it is also hardly readable for humans. So the big question is: \textbf{Why should we transform our colors to the \emph{HSV} space? Does it derive a benefit?}

The answer is yes. It is hardly readable for humans but it is way better to filter for specific colors. If we look at the definition openCV gives for the \emph{RGB} space, the higher complexity for some tasks becomes obvious:
\begin{quote}
In the \emph{RGB} color space all "the three channels are effectively correlated by the amount of light hitting the surface", so the color and light properties are simply not separated, \cite{GUPTA2017}.
\end{quote}
Expressed in a simpler way: In the \emph{RGB} space the colors also influence the brightness and the brightness influences the colors. However, in the \emph{HSV} space, there is only one channel - the H channel - to describe the color. The S channel represents the saturation and H the intensity. This is the reason why it is super useful for specific color filtering tasks.

\begin{figure}[ht]
\centering
  \includegraphics[width=0.8\linewidth]{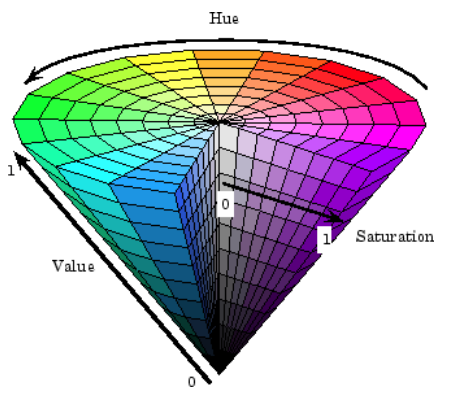}
  \caption{Illustration of the HSV Color Space, \cite{2018}}
  \label{fig:hsv_illustration}
\end{figure}

The \emph{HSV} color space is therefore often used by people who try to select specific colors. It corresponds better to how we experience color. As we let the H (Hue) channel go from 0 to 1, the colors vary from red through yellow, green, cyan, blue, magenta and back to red. So we have red values at 0 as well as at 1. As we vary the S (saturation) from 0 to 1, the colors simply vary from unsaturated (more grey like) to fully saturated (no white component at all). Increasing the V (value), the colors just become brighter. This color space is illustrated in figure~\ref{fig:hsv_illustration}.
Most systems use the so called \emph{RGB} additive primary colors. The resulting mixtures can be very diverse. The variety of colors, called the gamut, can therefore be very large. Anyway, the relationship between the constituent amounts of red, green, and blue lights is unintuitive.

\subsubsection{Derivation}

\begin{figure}[ht]
  \includegraphics[width=1.0\linewidth]{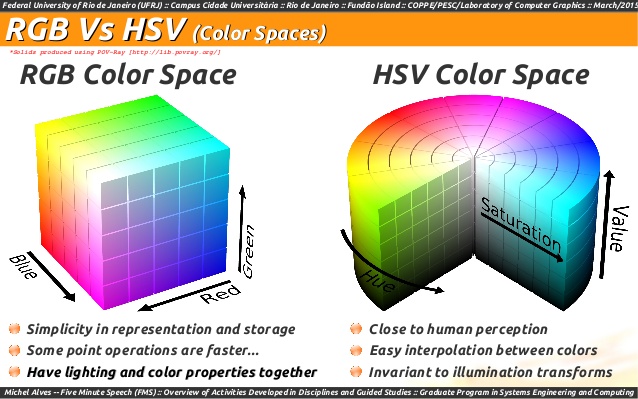}
  \caption{Comparison between the two Colors Spaces, \cite{Alves2015}}
  \label{fig:color_comparison}
\end{figure}

The \emph{HSV} model can be derived using geometric strategies. The \emph{RGB} color space is simply a cube where the addition of the three color components (with a scale form 0 to 1) is displayed. You can see this on the left of figure~\ref{fig:color_comparison}.

\begin{figure}[ht]
  \includegraphics[width=0.8\linewidth]{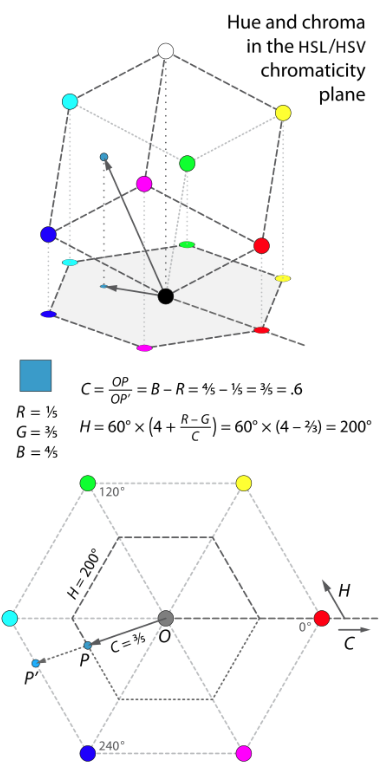}
  \caption{Cutting the Cube, \cite{2017a}}
  \label{fig:derivation_1}
\end{figure}

You can now simply take this cube and tilt it on its corner. We do it the way so that black rests at the origin whereas white is the highest point directly above it along the vertical axis. Afterwards you can just measure the hue of the colors by their angle around the vertical axis (red is denoted as 0$^{\circ}$). Going from the middle to the outer parts from 0 (where the grey like parts are) to 1 determines the saturation. This is illustrated in figure~\ref{fig:derivation_1}.

\begin{figure}[ht]
  \includegraphics[width=1.0\linewidth]{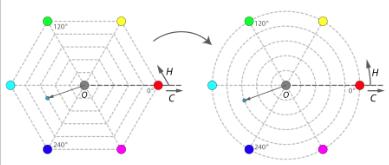}
  \caption{Warping Hexagons to Circles, \cite{2017a}}
  \label{fig:derivation_2}
\end{figure}

The definitions of hue and chroma (proportion of the distance from the origin to the edge of the hexagon) amount to a geometric warping of hexagons into circles. Each side of the hexagon is mapped linearly onto a 60$^{\circ}$) arc of the circle. This is visualized in figure~\ref{fig:derivation_2}.

For the value or lightness there are several possibilities to define an appropriate dimension for the color space. The simplest one is just the average of the three components, which is nothing else then the vertical height of a point in our tilted cubic. For this case we have:
\begin{equation}
V = 1/3 * (R + G + B)
\end{equation} 
For another definition, the value is defined as the largest component of a color. This places all three primaries and also all of the "secondary colors" (cyan, magenta, yellow) into a plane with white. This forms a hexagonal pyramid out of the RGB cube. This is called the HSV "hexcone" model and is the common one. We get:
\begin{equation}
V = max(R, G, B)
\end{equation} 

\subsubsection{In Practice}

\begin{enumerate}
\item Form a hexagon by projecting the RGB unit cube along its pincipal diagonal onto a plane (figure~\ref{fig:hsv_practice_1}).
\begin{figure}[ht]
  \includegraphics[width=1.0\linewidth]{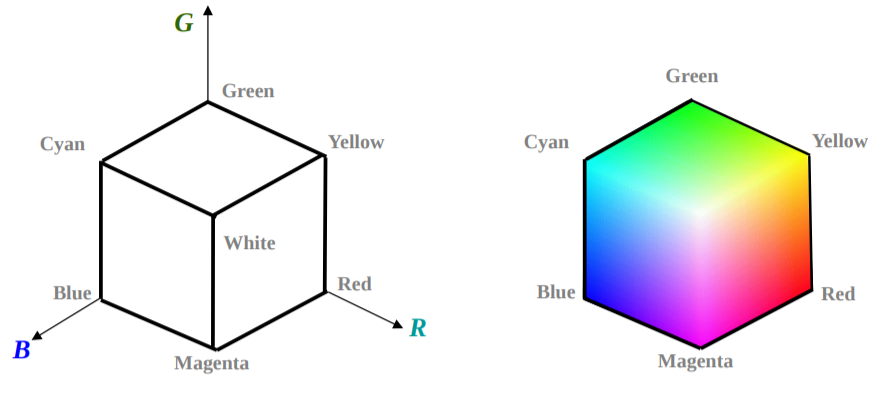}
  \caption{First layer of the Cube (left) and Flat Hexagon (right) \cite{CHENG}}
  \label{fig:hsv_practice_1}
\end{figure}
\item Repeat projection with smaller RGB cube (subtract 1/255 in length of every cube) to obtain smaller projected hexagon. Like this a HSV hexcone is formed by stacking up the 256 hexagons in decreasing order of size (figure~\ref{fig:hsv_practice_2}).
\begin{figure}[ht]
  \includegraphics[width=1.0\linewidth]{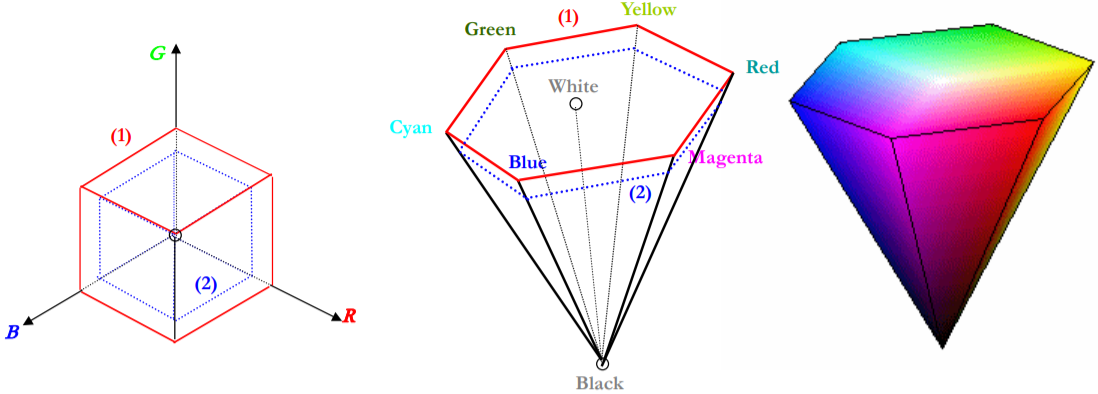}
  \caption{Stacking hexagons together \cite{CHENG}}
  \label{fig:hsv_practice_2}
\end{figure}
Then the value is again defined as: \newline
$V = max(R, G, B)$
\item Smooth edges of hexagon to circles (see previous chapter).
\end{enumerate}

\subsubsection{Application}

\begin{figure}[ht]
  \includegraphics[width=1.0\linewidth]{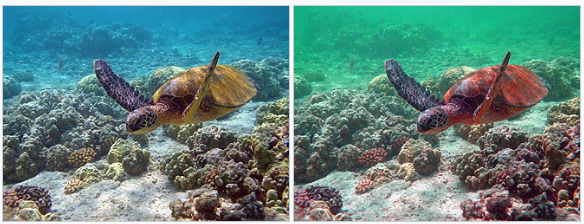}
  \caption{Image on the left is original. Image on the right was simply produced by rotating the H of each color by -30$^{\circ}$ while keeping S and V constant \cite{2017a}}
  \label{fig:hsv_example}
\end{figure}

One nice example of the application of the HSV color space can be seen in figure~\ref{fig:hsv_example}.

It just shows how simple color manipulation can be performed in a very intuitive way. We can turn many different applications to good account using this approach. As you have seen, color filtering also simply becomes a threshold query.

\fontsize{9.0pt}{10.0pt}
\selectfont
\bibliographystyle{aaai}
\bibliography{saviors}

\end{document}